\documentclass[review]{elsarticle}

\usepackage{lineno,hyperref,subcaption}
\usepackage{algorithm}
\usepackage[noend]{algpseudocode}
\modulolinenumbers[5]
\usepackage{amsmath}

\journal{}

\begin{document}

\begin{frontmatter}

\title{Deep EndoVO: A Recurrent Convolutional Neural Network (RCNN) based Visual Odometry Approach for Endoscopic Capsule Robots}
%\tnotetext[mytitlenote]{Fully documented templates are available in the elsarticle package on \href{http://www.ctan.org/tex-archive/macros/latex/contrib/elsarticle}{CTAN}.}

%% Group authors per affiliation:
%\author{Elsevier\fnref{myfootnote}}
%\address{Radarweg 29, Amsterdam}
%\fntext[myfootnote]{Since 1880.}

%% or include affiliations in footnotes:
\author[address1,address3]{Mehmet Turan}
\ead{mturan@student.ethz.ch}
\author[address2]{Yasin Almalioglu}
\ead{yasin.almalioglu@boun.edu.tr}
\author[address22]{Helder Araujo}
\ead{helder@isr.uc.pt}
\author[address3]{Ender Konukoglu}
\ead{ender.konukoglu@vision.ee.ethz.ch}
\author[address1]{Metin Sitti}
\ead{sitti@is.mpg.de}
\address[address1]{Physical Intelligence Department, Max Planck Institute for Intelligent Systems, Germany}
\address[address2]{Computer Engineering Department, Bogazici Univesity, Turkey}
\address[address22]{Institute for Systems and Robotics, Universidade de Coimbra,Portugal}
\address[address3]{ Computer Vision Laboratory, Department of Information Technology and Electrical Engineering, ETH Zurich, Switzerland}

\begin{abstract}
	
Ingestible wireless capsule endoscopy is an emerging  minimally invasive diagnostic technology for inspection of the GI  tract and diagnosis of a wide range of diseases and pathologies. Medical device companies and many research groups have recently made substantial progresses in converting passive capsule endoscopes to active capsule robots, enabling more accurate, precise, and intuitive detection of the location and size of the diseased areas. Since a reliable real time pose estimation functionality is crucial for actively controlled endoscopic capsule robots, in this study,  we propose a monocular visual odometry (VO) method for endoscopic capsule robot operations. Our method lies on the application of the deep Recurrent Convolutional Neural Networks (RCNNs) for the visual odometry task, where Convolutional Neural Networks (CNNs) and Recurrent Neural Networks (RNNs)  are used for the feature extraction and inference of dynamics across the frames, respectively. Detailed analyses and evaluations made on a real pig stomach dataset proves that our system achieves high translational and rotational accuracies for different types of endoscopic capsule robot trajectories.	
	
\end{abstract}


\end{frontmatter}

%\linenumbers

\section{Introduction}
\begin{figure}
	% Use the relevant command to insert your figure file.
	% For example, with the graphicx package use
	\includegraphics[width=\textwidth]{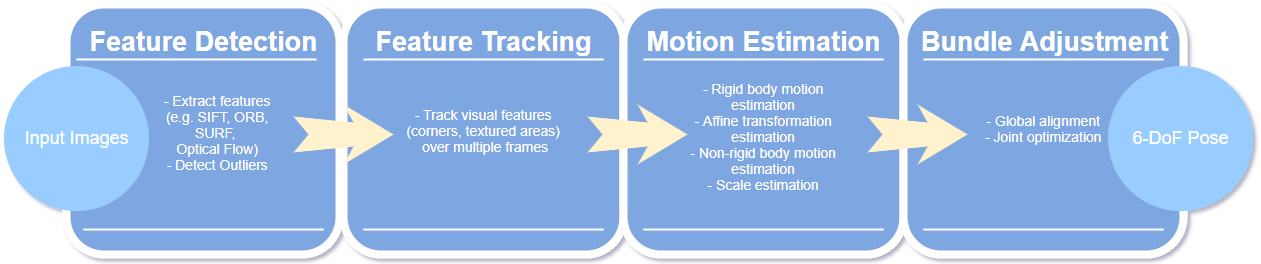}
	% figure caption is below the figure
	\caption{Traditional visual odometry pipeline.}
	\label{fig:trad}       % Give a unique label
\end{figure}

 Following the advances in material science in last decades, untethered pill-size, swallowable capsule endoscopes with an on-board camera and wireless image transmission device have been developed and used in hospitals for screening the gastrointestinal tract and diagnosing diseases such as the inflammatory bowel disease, the ulcerative colitis and the colorectal cancer. Unlike standard endoscopy, endoscopic capsule robots are non-invasive, painless and more appropriate to be employed for long duration screening purposes. Moreover, they can access difficult body parts that were not possible to reach before with standard endoscopy (e.g., small intestines). Such advantages make pill-size capsule endoscopes a significant alternative screening method over standard endoscopy \citep{liao2010indications, nakamura2008capsule, pan2011swallowable, than2012review, sitti2015biomedical}. However, current capsule endoscopes used in hospitals are passive devices controlled by peristaltic motions of the inner organs. The control over capsule's position, orientation, and functions would give the doctor a more precise reachability of targeted body parts and more intuitive and correct diagnosis opportunity \citep{turan2017non, turan2017deep, turan2017six, turan2017fully, turan2017sparse}. Therefore, several groups have recently proposed active, remotely controllable robotic capsule endoscope prototypes equipped with additional functionalities such as local drug delivery, biopsy and other medical functions \citep{goenka2014capsule, nakamura2008capsule, munoz2014review, carpi2011magnetically, keller2012method, mahoney2013managing, yim2014biopsy, petruska2013omnidirectional, son20165, son2017magnetically}. However, an active motion control needs feedback from a precise and reliable real time pose estimation functionality. In last decade, several localization methods \citep{than2012review, fluckiger2007ultrasound, rubin2006sonographic, kim2008noninvasive, yim20133} were proposed to calculate the 3D position and orientation of the endoscopic capsule robot such as fluoroscopy \citep{than2012review}, ultrasonic imaging \citep{fluckiger2007ultrasound, rubin2006sonographic, kim2008noninvasive, yim20133}, positron emission tomography (PET) \citep{than2012review, yim20133}, magnetic resonance imaging (MRI) \citep{than2012review}, radio transmitter based techniques and magnetic field based techniques \citep{yim2014biopsy}. The common drawback of these localization methods is that they require extra sensors and hardware design. Such extra sensors have their own deficiencies and limitations if it comes to their application in small scale medical devices such as space limitations, cost aspects, design incompatibilities, biocompatibility issue and the interference of sensors with activation system of the device. \\
As a solution of these issues, a trend of visual odometry methods have attracted the attention for the localization of such small scale medical devices. A classic visual odometry pipeline   typically consisting of camera calibration, feature detection, feature matching, outliers rejection (e.g RANSAC), motion estimation, scale estimation and global optimization (bundle adjustment) is depicted in Fig. \ref{fig:trad}. 
Although some state-of-the-art algorithms based on this traditional pipeline have been applied for the visual odometry task of the hand-held endoscopes in the past decades, their main deficiency is tracking failures in low textured areas. In last years, deep learning (DL) techniques have been dominating many computer vision related tasks with some promising result, e.g object detection, object recognition, classification problems etc. Contrary to these high-level computer vision tasks, VO is mainly working on motion dynamics and relations across sequence of images, which can be defined as a sequential learning problem. With that motivation, we propose a novel monocular VO algorithm based on deep Recurrent Convolutional Neural Networks (RCNNs). Since it is designed in an end-to-end fashion, it does not need any module from the classic VO pipeline to be integrated. The main contributions of our paper are as follows: 
\begin{itemize}
	\item To the best of our knowledge, this is the first  monocular VO approach through deep learning techniques developed for the endoscopic capsule robot and hand-held standard endoscope localization.
	\item Neither prior knowledge nor parameter tuning is needed to recover the absolute trajectory scale contrary to monocular traditional VO approach.
	\item A novel RCNN architecture is introduced which can successfully  model sequential dependence and complex motion dynamics across endoscopic video frames.
	\item A real pig stomach dataset and a synthetic human simulator dataset with 6-DoF ground truth pose labels and 3D scan are recorded, which we are considering to publish for the sake of other researchers in that area.
\end{itemize}
The proposed method solves several issues faced by typical visual odometry pipelines, e.g the need to establish a frame-to-frame feature correspondence, vignetting, motion blur, specularity or low signal-to-noise ratio (SNR). We think that DL based endoscopic VO approach is more suitable for such challenge areas since the operation environment (GI tract) has similar organ tissue patterns among different patients which can be learned by a sophisticated machine learning approach easily. Even the dynamics of common artefacts such as vignetting, motion blur and specularity across frame sequences could be learned and used for a better pose estimation.\\
As the outline of this paper, Section \ref{sec:alg_analysis} introduces the proposed RCNN based localization method in detail. Section \ref{sec:dataset}  presents our dataset and the experimental setup. Section \ref{sec:results} shows our experimental results, we achieved for 6-DoF localization of the endoscopic capsule robot. Section \ref{sec:conclusion} gives future directions.

\begin{figure}
	% Use the relevant command to insert your figure file.
	% For example, with the graphicx package use
	\includegraphics[width=\textwidth]{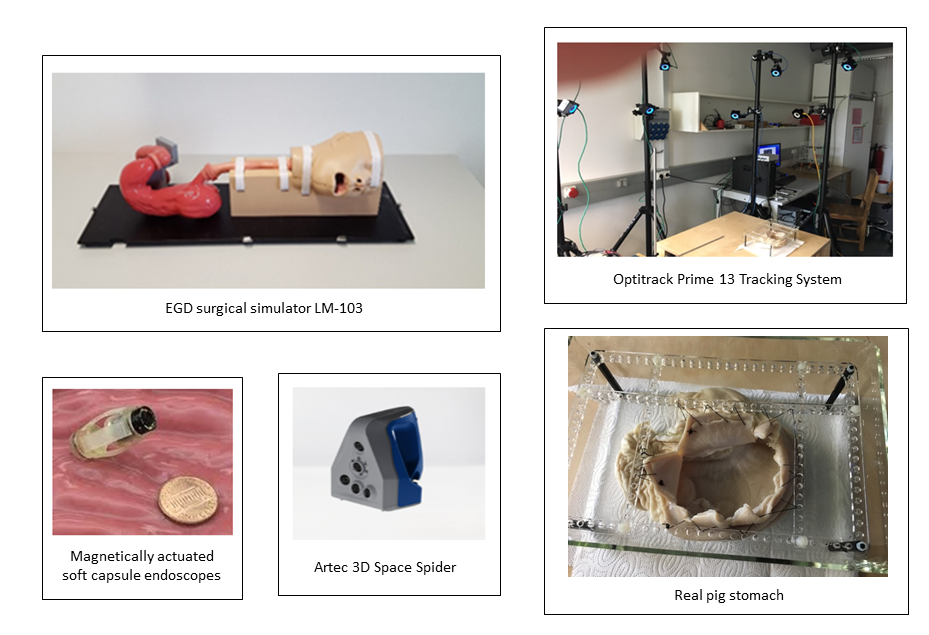}
	% figure caption is below the figure
	\caption{Experimental overview.}
	\label{fig:architecture}       % Give a unique label
\end{figure}

\section{System Overview and Analysis} \label{sec:alg_analysis}
Our architecture makes use of inception modules for feature extraction and RNN for sequential modelling of motion dynamics to regress the robot's orientation and position in real time (5.3 ms per frame). It takes two consecutive endoscopic RGB Depth frames each with timestamp and regresses the 6-DoF pose of the robot without need of any extra sensor. For the depth image creation from RGB input images, we used shape from shading (SfS) technique of Tsai and Shah, which is based on the following assumptions \citep{PingSing1994}:
\begin{itemize}
	\item	The object surface is lambertian
	\item	The light comes from a single point light source
	\item	The surface has no self-shaded areas. 
\end{itemize}
For more details of the Tsai-Shah SfS method, the reader is referred to the original paper of the authors. In past couple of years, some powerful CNN architectures, such as GoogleNet\citep{szegedy2015going}, VGG16\citep{Simonyan14c}, ResNet50\citep{he2016deep}  have been developed and evaluated for various high level computer vision tasks, e.g object detection, object recognition and classification \citep{szegedy2015going},\citep{sunderhauf2015performance},\citep{russakovsky2015imagenet}\citep{kendall2015posenet}. One major drawback of CNN architectures is the fact that they only analyse just-in-moment information, whereas VO is rather dependent on the correlative information across frames. Unlike traditional feed-forward artificial neural networks, RCNN can use its internal memory to process arbitrarily long sequences by its directed cycles between the hidden units.  Therefore, we think that RCNN architectures are more suitable than CNN architectures for VO tasks. The proposed deep EndoVO (endoscopic visual odometry) approach works as follows: 

\begin{algorithm}
		\caption{Deep EndoVO}
	\label{algo:EndoVO}
	\begin{algorithmic}[1]
		\State Take two consecutive input RGB images.
		\State Create the depth images from RGB images using Tsai-Shah SfS method.
		\State Subtract mean RGB Depth value of the training set from the RGB Depth images.
		\State Stack the preprocessed RGB Depth frame pair to form a tensor.
		\State Serve the tensor into the stack of inception modules to create the feature vector.
		\State Feed the feature representation into the RNN layers.
		\State Estimate the 6-DoF relative pose. 
	\end{algorithmic}
\end{algorithm}

%\begin{figure}
%  \includegraphics[width=\textwidth]{figures/figset.png}
%\caption{Architecture Block Diagram.}
%\label{fig:architectured}       % Give a unique label
%\end{figure}

\begin{figure}[t!]
\begin{subfigure}[t]{0.5\textwidth} 
\includegraphics[width=\textwidth]{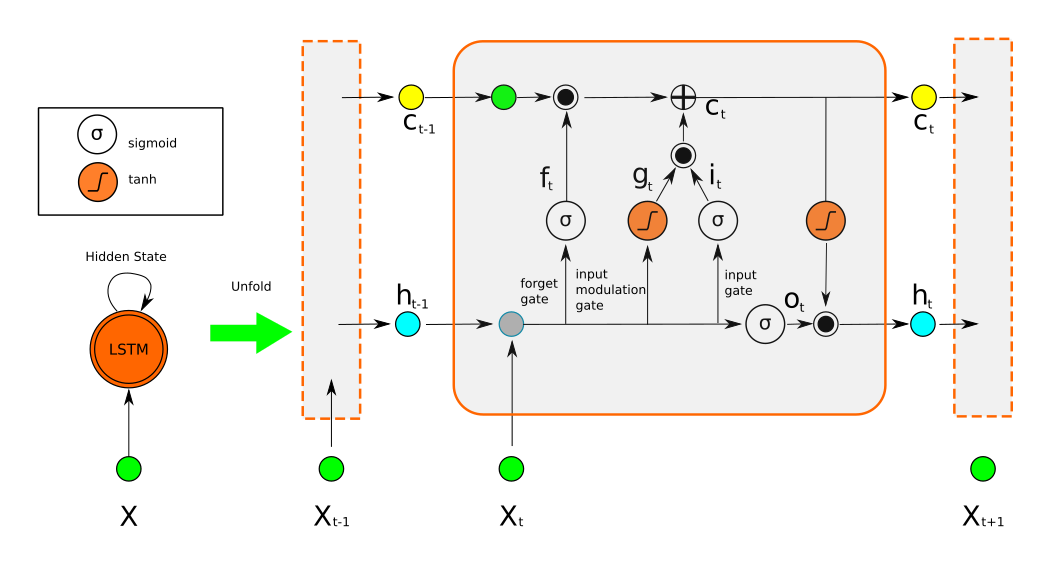}
	% figure caption is below the figure
	\caption{ Information flow through the hidden units of the LSTM \citep{gers1999learning}.}
	\label{fig:RNN}       % Give a unique label
\end{subfigure}
%\\
~
\begin{subfigure}[t]{0.5\textwidth} 
\includegraphics[width=\textwidth]{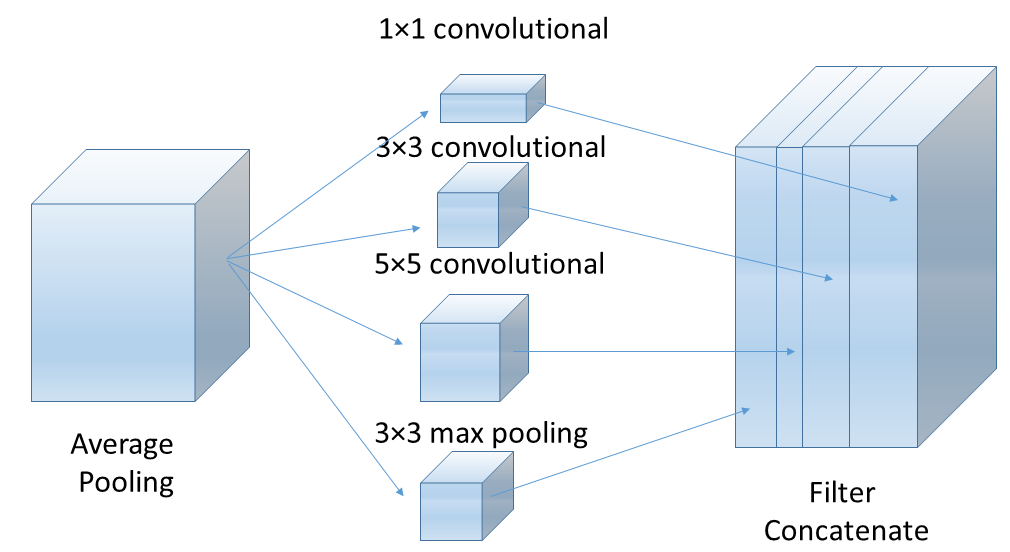}
	% figure caption is below the figure
	\caption{Inception layer\citep{szegedy2015going}}
	\label{fig:incept}       % Give a unique label
\end{subfigure}
\caption{The structure of the LSTM and inception layers of the proposed model is shown.}
\label{fig:deep_modules}
\end{figure}
The proposed DL network consists of three inception layers and two LSTM layers concatenated sequentially. The inception layers, imitating  visual cortex of human beings, are basically extracting multi-level features; i.e,  features of different sizes such as small details, middle-size or larger features (see Fig. \ref{fig:incept}. The final inception layer passes the feature representation into the RNN modules (see Fig. \ref{fig:RNN}). RNNs are very suitable for modelling the dependencies across image sequences and for creating a temporal motion model since it has a  memory of hidden states over time and has directed cycles among hidden units, enabling the current hidden state to be a function of arbitrary sequences of inputs (see Fig. \ref{fig:RNN}). Thus, using RNN, the pose estimation of the current frame benefits from information encapsulated in previous frames \citep{walch2016image, wang2017deepvo}. Given a set of inception features $x_{k}$ at time $k$, RNN updates at time step $k$, $W$ denote corresponding weight matrices of the hidden units, $b$ the bias vector, and $H$  an element-wise hyperbolic tangent based activation function. Long Short-Term Memory (LSTM) is more suitable than RNN to exploit longer trajectories since it avoids the vanishing gradient problem of RNN resulting in a higher capacity of learning long-term relations among the sequences by introducing memory gates such as input, forget and output gates and hidden units of several blocks. The input gate controls the amount of new information flowing into the current state, the forget gate adjusts the amount of existing information that remains in the memory and the output gate decides which part of the information triggers  the activations. The folded LSTM and its unfolded version over time are shown in Fig. \ref{fig:RNN} along with the internal structure of a LSTM memory cell. It can be seen that unfolded LSTMs correspond to timestamps. Given the input vector $x_{k}$ at time $k$, the output vector $h_{k-1}$ and the cell state vector $c_{k-1}$ of the previous LSTM unit, the LSTM updates at time step $k$ according to the following equations, where $\sigma$ is sigmoid non-linearity, tanh is hyperbolic tangent non-linearity, $W$ terms denote corresponding weight matrices, $b$ terms denote bias vectors, $i_{k}$, $f_{k}$, $g_{k}$, $c_{k}$ and $o_{k}$ are input gate, forget gate, input modulation gate, the cell state and output gate at time $k$, respectively \citep{gers1999learning}:

%\begin{equation}

\begin{align*}
f_k &= \sigma(W_{f}\cdot [x_k, h_{k-1}] + b_f) &
i_k &= \sigma(W_{i}\cdot [x_k, h_{k-1}] + b_i) \\
g_k &= tanh(W_{g}\cdot[x_k, h_{k-1}] + b_g) &
c_k &= f_k\odot c_{k-1} + i_k\odot g_k \\
o_k &= \sigma(W_{o}|cdot[x_k, h_{k-1}] + b_o) &
h_k &= o_k\odot tanh(c_k) 
\end{align*}

\begin{figure}
	% Use the relevant command to insert your figure file.
	% For example, with the graphicx package use
	\includegraphics[width=\textwidth]{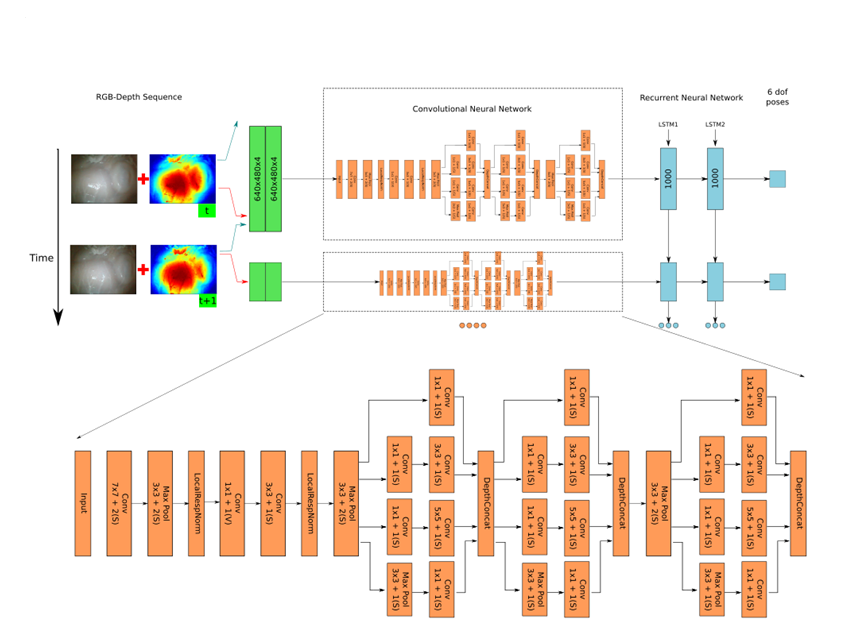}
	% figure caption is below the figure
	\caption{Architecture of the proposed RCNN based monocular VO system.}
	\label{fig:CNN}       % Give a unique label
\end{figure}

%\end{equation}
Although the LSTM is prone to vanishing gradient problem of RNN and is capable to detect the long-term dependencies, its learning capacity can be increased further by stacking multiple LSTM layers vertically. Thus, our deep RNN consists of two LSTM layers with the output sequence of the first one forming the input sequence of the second one each containing $1000$ hidden units, as illustrated in Fig. \ref{fig:CNN}. The proposed system, which learns translational and rotational motions simultaneously to regress the 6-DoF pose, is trained on Euclidean loss using Adam optimization method with the following objective loss function:
\begin{equation} \label{eqn:loss}
loss(I) = \|\hat{x} - x\|_2 + \beta\|\hat{q} - q\|_2
\end{equation}
where $x$ is the translation vector and $q$ is the rotation vector. The pseudo-code to calculate the loss value is given in Algorithm \ref{algo:calc_loss}. In our loss function, a balance $\beta$ must be kept between the orientation and translation loss values which are highly coupled each other as they are learned from the same model weights. Experimental results show that the optimal $\beta$ is given by the ratio between the loss values of predicted positions and orientations at the end of training session \citep{kendall2015posenet}.

\begin{algorithm}
\caption{Pseudo code to calculate the loss over the network}
\label{algo:calc_loss}
\begin{algorithmic}[1]
	\Procedure{CalculateLoss}{}
	\State $loss$ $\gets 0$
	\For{$layer$ in $layers$}
	\For{$top, loss\_weight$ in $layer.tops, layer.loss\_weights$}
		\State $loss \gets loss+loss\_weight \times sum(top)$
	\EndFor
	\EndFor
	\EndProcedure
\end{algorithmic}
\end{algorithm}
The back-propagation algorithm is used to calculate the gradients of RCNN weights, which are passed to the Adam optimization method to compute adaptive learning rates for each parameter employing the first-order gradient-based optimization of the stochastic objective function. In addition to saving exponentially decaying average of past squared gradients, $v_t$, Adam optimization keeps exponentially decaying average of past gradients, $m_t$ that is similar to momentum.  The update equations are given as 
\begin{equation} \label{eqn:weight_update1}
(m_t)_i = \beta_1 (m_{t-1})_i + (1-\beta_1 )(\nabla L(W_t))_{i}
\end{equation}
\begin{equation} \label{eqn:weight_update2}
(v_t)_i = \beta_2 (v_{t-1})_i + (1-\beta_2)(\nabla L(W_t))_i^2
\end{equation}
\begin{equation} \label{eqn:weight_update3}
(W_{t+1})_i = (W_t)_i - \alpha \frac{\sqrt{1-(\beta_2)_i^t}}{1- (\beta_1)_i^t} \frac{(m_t)_i}{\sqrt{(v_t)_i+\varepsilon}}
\end{equation}
We used default values proposed  by \citep{kingma2014adam} for the parameters $\beta_1,  \beta_1$ and $\varepsilon$: $\beta_1=0.9$, $\beta_2=0.999$ and $\varepsilon=10^{-8}$. 

%
%\subsection{Transfer Learning}
%We sidestep the common problem of  that it requires large amount of training data by using transfer learning for the first layers of the proposed architecture. For that aim, we started the pose training using the weights acquired by ImageNet \citep{wugpu}. According to the experimental results, we were able to conclude that a network trained to output pose-invariant classification labels is suitable as a starting point for a pose regressor in the context of endoscopic capsule robot localization. A possible explanation of that observation is that the network must have activation units that are invariant to different pose values. During transfer learning on the first layers, we kept the learning rate low in order not to distort the already learned weights. During the training of next layers, the learning rate of individual models were increased to enable a learning procedure from scratch. We aim to publish the weights of the trained model for the sake of other research groups working on endoscopic capsule robot localization.

\section{Dataset} \label{sec:dataset}
This section demonstrates the experimental setup of the proposed study, introduces our magnetically actuated soft capsule endoscopes (MASCE) and explains how the  training and testing datasets were recorded. 

\subsection{Magnetically Actuated Soft Capsule Endoscopes (MASCE)}
\begin{figure}[t!]
\begin{subfigure}[t]{0.6\textwidth}
\centering
	\includegraphics[width=0.6\textwidth]{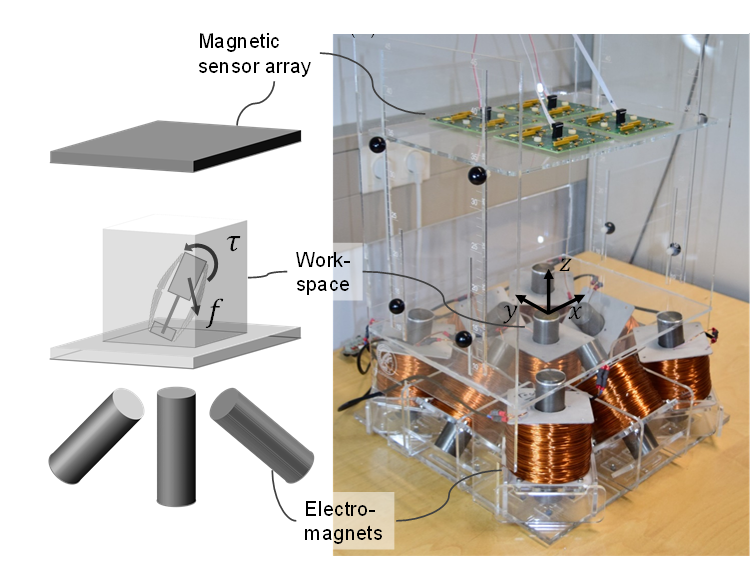}
	% figure caption is below the figure
	\caption{Actuation system of the MASCE \citep{son2017magnetically} }
	\label{fig:07r}       % Give a unique label
\end{subfigure}
%\\
\bigskip
\begin{subfigure}[t]{0.4\textwidth} 
\includegraphics[width=\textwidth]{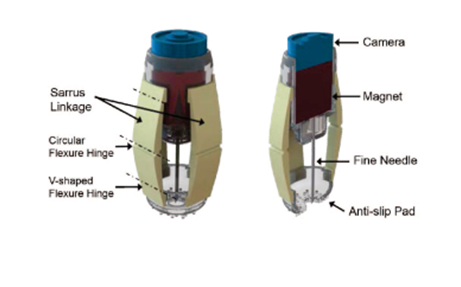}
	% figure caption is below the figure
	\caption{Exterior (left) and section view  (right) of MASCE \citep{son2017magnetically}}
	\label{fig:magn_cross}       % Give a unique label
\end{subfigure}
\caption{MASCE design features and actuation unit}
\label{fig:capsule_setup}
\end{figure}

Our capsule prototype is a magnetically actuated soft capsule endoscope (MASCE) designed for disease detection, drug delivery and biopsy operations in the upper gastrointestinal tract. The prototype is composed of a RGB camera, a permanent magnet, a fine-needle and a drug chamber (see Fig. \ref{fig:capsule_setup} for visual reference). The magnet exerts magnetic force and torque to the robot in response to a controlled external magnetic field \citep{son2017magnetically}. The magnetic torque and forces are used to actuate the capsule robot and to release drug and deliver the needle through the hole in the bottom of the capsule. Magnetic fields from the electromagnets generate the magnetic force and torque on the magnet inside MASCE so that the robot moves inside the workspace. Sixty-four three-axis magnetic sensors are placed on the top, and nine electromagnets are placed in the bottom \citep{son2017magnetically}.

\subsection{Training dataset} \label{sec:traindataset_equip}
\begin{figure}[t!]
\begin{subfigure}[t]{0.5\textwidth} 
\includegraphics[width=\textwidth]{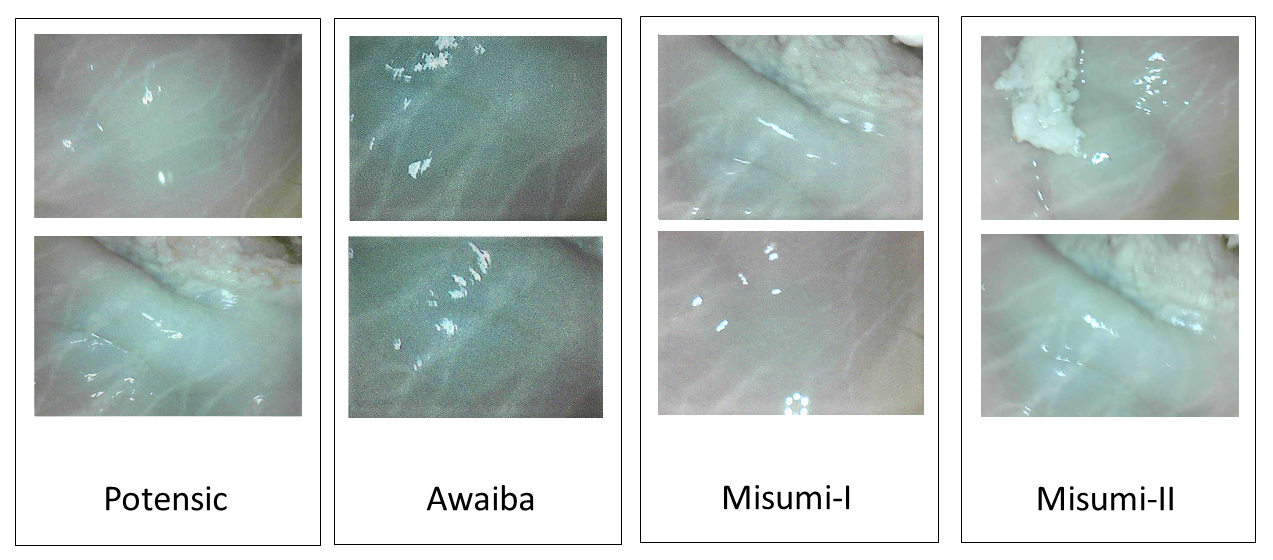}
% figure caption is below the figure
\caption{Sample frames recorded on a real pig stomach}
\label{fig:07}
\end{subfigure}
%\\
~
\begin{subfigure}[t]{0.5\textwidth} 
\includegraphics[width=\textwidth]{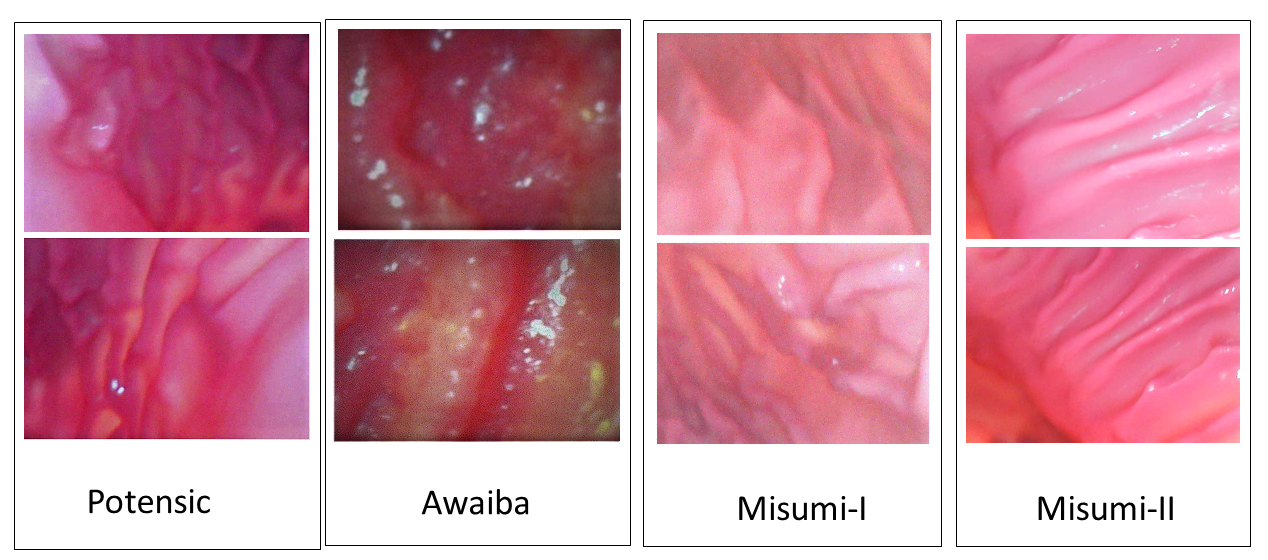}
% figure caption is below the figure
\caption{Sample frames recorded on EGD simulator }
\label{fig:037}       % Give a unique label 
\end{subfigure}
\caption{Sample frames from the datasets used in the experiments.}
\label{fig:loss}
\end{figure}

We created two groups of training datasets. The first training dataset was recorded on five different real pig stomachs (see Fig.\ref{fig:architecture}), whereby the second dataset which was only used for training purposes, was captured using a non-rigid open GI tract model EGD (esophagus gastro duodenoscopy) surgical simulator LM-103 (see Fig.\ref{fig:architecture}). To ensure that our algorithm is not tuned to a specific camera model, four different commercial endoscopic cameras were employed, specifications of which are shown in Table \ref{tab:cam_specs}, accordingly. For each pig stomach-camera combination, $2000$ frames were acquired  which makes for four cameras and five pig stomachs $40000$ frames, in total.  Sample real pig stomach frames are shown in Fig. \ref{fig:07} for visual reference. As a second training dataset, for each of four cameras, we captured $10000$ frames on an EGD human stomach simulator making $40000$ frames, in total. Sample synthetic training frames are shown in Fig.\ref{fig:037} for visual reference. During video recording, Optitrack motion tracking system consisting of eight Prime-13 cameras and a tracking software was utilized to obtain 6-DoF localization ground truth data in a sub-millimeter precision (see Fig. \ref{fig:architecture}) which was used as a gold standard for the evaluations of the pose estimation accuracy.
 
 \begin{table}[t!]
\begin{subtable}[t]{.5\textwidth}
	\centering
	% table caption is above the table
	\caption{Awaiba Naneye Endoscopic Camera }
	\label{tab:1aa}       % Give a unique label
	%
	% For LaTeX tables use
	\begin{tabular}{ccccc}
		\hline\noalign{\smallskip}
		Resolution & 250 x 250 pixel \\
		\hline\noalign{\smallskip}
		Footprint & 2.2 x 1.0 x 1.7 mm \\
		\hline\noalign{\smallskip}
		Pixel size & 3 x 3 $\mu m^2$\\
		\hline\noalign{\smallskip}
		Frame rate& 44 fps \\
		\hline\noalign{\smallskip}
	\end{tabular}
\end{subtable}
~
\begin{subtable}[t]{.5\textwidth}
	\centering
	% table caption is above the table
	\caption{Misumi-V3506-2ES camera }
	\label{tab:1ab}       % Give a unique label
	%
	% For LaTeX tables use
	\begin{tabular}{ccccc}
		\hline\noalign{\smallskip}
		Resolution & 400 x 400 pixel \\
		\hline\noalign{\smallskip}
		Diameter & 8.2mm \\
		\hline\noalign{\smallskip}
		Pixel size & 5.55 x 5.55 $\mu m^2$ \\
		\hline\noalign{\smallskip}
		Frame rate& 30 fps\\
		\hline\noalign{\smallskip}
	\end{tabular}
\end{subtable}
~
\begin{subtable}[t]{.5\textwidth}
	\centering
	% table caption is above the table
	\caption{Misumi-V5506-2ES camera }
	\label{tab:1ac}       % Give a unique label
	%
	% For LaTeX tables use
	\begin{tabular}{ccccc}
		\hline\noalign{\smallskip}
		Resolution & 640 x 480 pixel \\
		\hline\noalign{\smallskip}
		Diameter & 8.6 mm \\
		\hline\noalign{\smallskip}
		Pixel size & 6.0 x 6.0 $\mu m^2$ \\
		\hline\noalign{\smallskip}
		Frame rate& 30 fps \\
		\hline\noalign{\smallskip}
	\end{tabular}
\end{subtable}
~
\begin{subtable}[t]{.5\textwidth}
	\centering
	% table caption is above the table
	\caption{Potensic Mini Camera}
	\label{tab:1acd}       % Give a unique label
	%
	% For LaTeX tables use
	\begin{tabular}{ccccc}
		\hline\noalign{\smallskip}
		Resolution & 1280 x 720 pixel \\
		\hline\noalign{\smallskip}
		Diameter & 8.8 mm \\
		\hline\noalign{\smallskip}
		Pixel size & 10.0 x 10.0 $\mu m^2$ \\
		\hline\noalign{\smallskip}
		Frame rate& 30 fps \\
		\hline\noalign{\smallskip}
	\end{tabular}
\end{subtable}
\caption{Endoscopic camera specifications used for the experiments.}
\label{tab:cam_specs}
\end{table}

\subsection{Testing dataset} \label{sec:testdataset_equip}
 We created a testing dataset recorded using five different real pig stomachs, which were not used for the training section. For each pig stomach-camera combination, $2000$ frames are acquired making $40000$ frames, in total. We did not capture any synthetic dataset for the testing session since it is less realistic due to obvious patterns of such artificial simulators. For all of the video records, again Optitrack motion tracking system  was utilized to obtain 6-DoF localization ground truth.

\section{Evaluations and Results}
\label{sec:results}

\begin{figure}[t!]
\begin{subfigure}[t]{0.5\textwidth} 
\includegraphics[width=\textwidth]{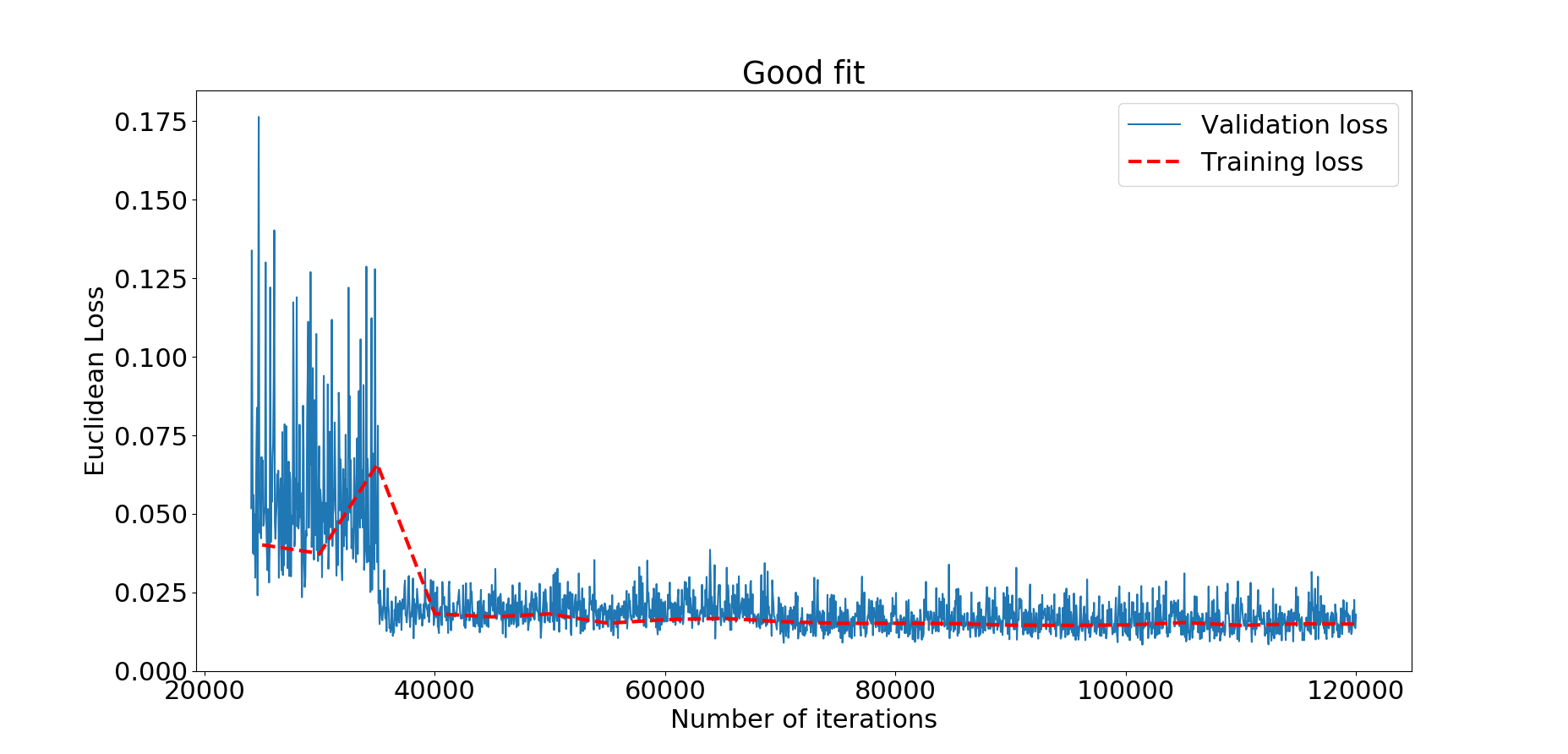}
\caption{Change in the loss values for a good fitting}
\label{fig:loss_goodfit}       % Give a unique label
\end{subfigure}
%\\
~
\begin{subfigure}[t]{0.5\textwidth} 
\includegraphics[width=\textwidth]{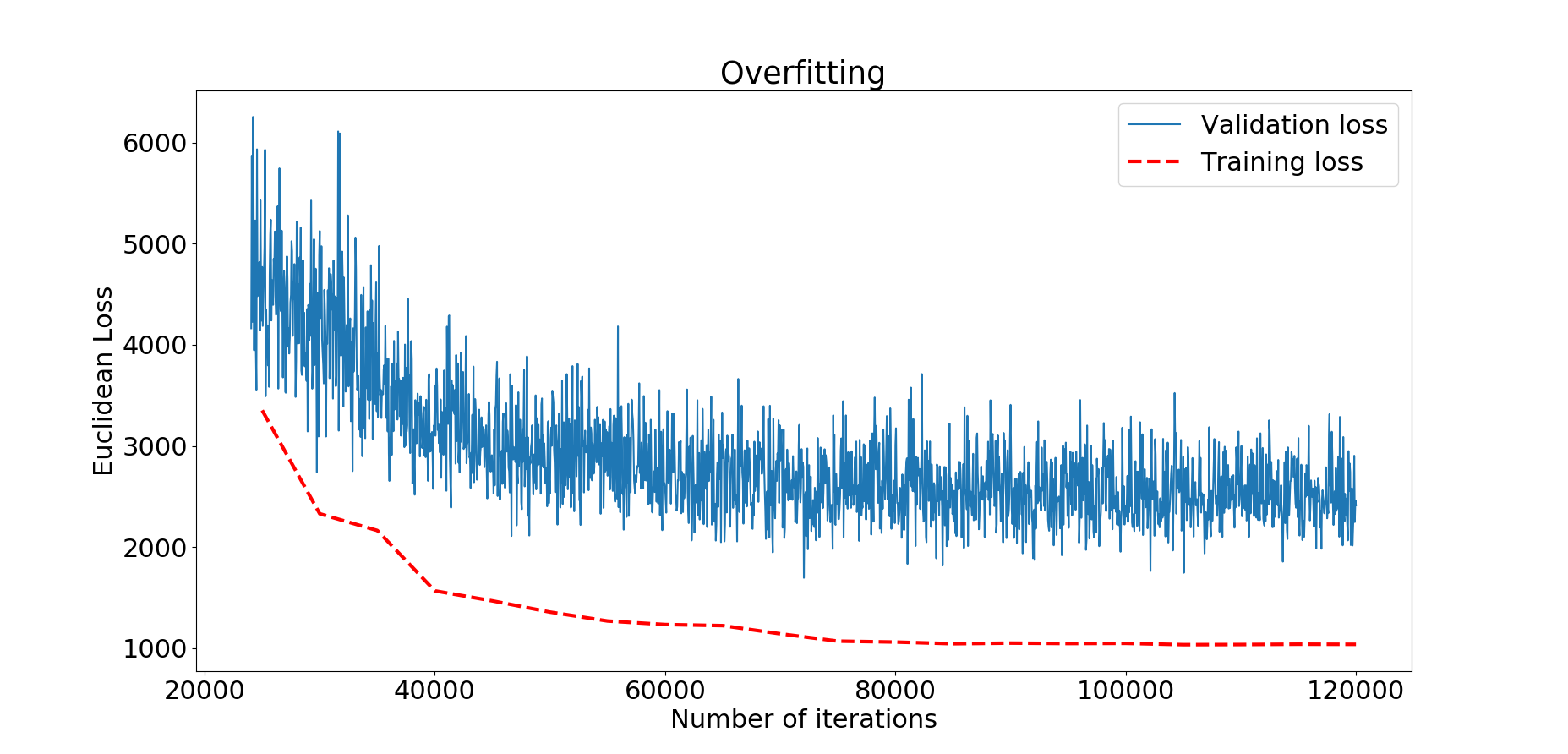}
\caption{Change in the loss values for overfitting}
\label{fig:loss_overfit}
\end{subfigure}
\caption{The decrease in the training and validation loss values. In overfitting case, the training loss gets smaller than the validation loss. However, the loss values are balanced for a good fit.}
\label{fig:loss}
\end{figure}

\begin{figure}
\begin{subfigure}{0.5\textwidth} 
\includegraphics[width=\textwidth]{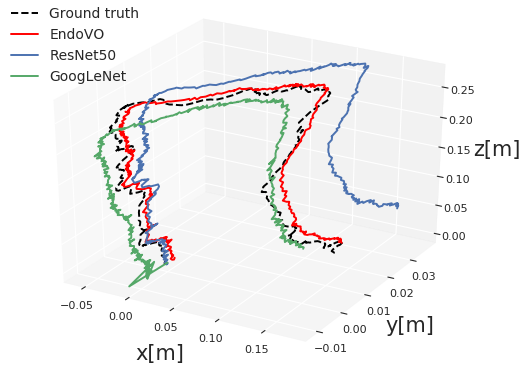}
\caption{Trajectory 1}
\label{fig:traj_1}       % Give a unique label
\end{subfigure}
%~
\begin{subfigure}{0.5\textwidth} 
\includegraphics[width=\textwidth]{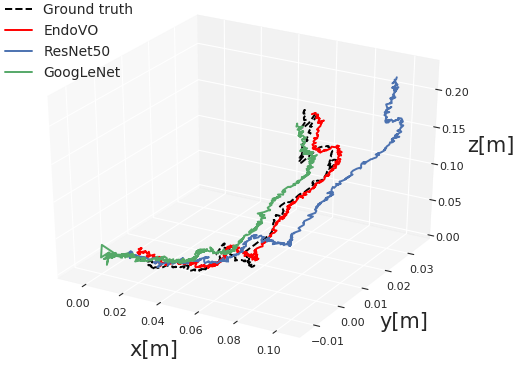}
\caption{Trajectory 2}
\label{fig:traj_2} 
\end{subfigure}
%\\
\begin{subfigure}{0.5\textwidth} 
\includegraphics[width=\textwidth]{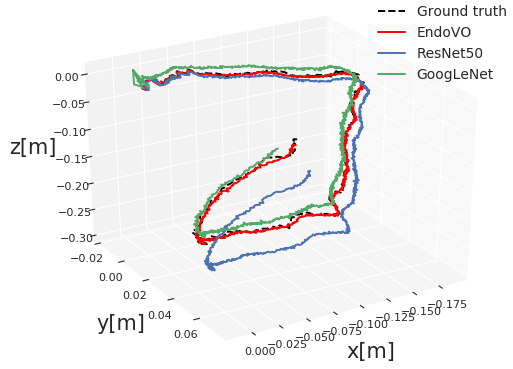}
\caption{Trajectory 3}
\label{fig:traj_3} 
\end{subfigure}
%~
\begin{subfigure}{0.5\textwidth} 
\includegraphics[width=\textwidth]{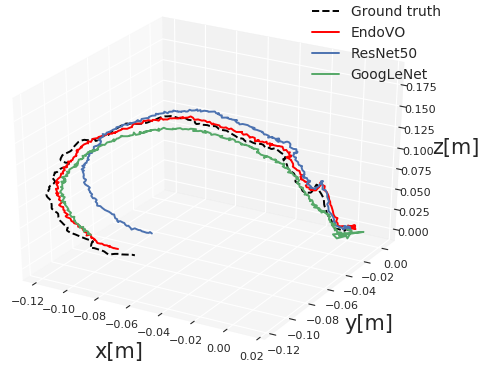}
\caption{Trajectory 4}
\label{fig:traj_4} 
\end{subfigure}
%~
\caption{Sample ground truth trajectories and estimated trajectories predicted by the DL based VO models. As seen, deep EndoVO is the closest to the ground truth trajectories. The scale is calculated and maintained correctly by the models.}
\label{fig:trajectories}
\end{figure}

\begin{figure}[t!]
\begin{subfigure}[t]{0.5\textwidth} 
\includegraphics[width=\textwidth]{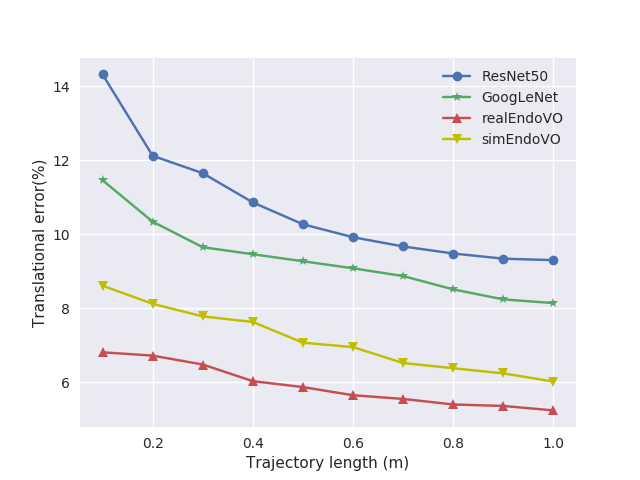}
\caption{Trajectory length vs translation error}
\label{fig:trans_error} 
\end{subfigure}
~
\begin{subfigure}[t]{0.5\textwidth} 
\includegraphics[width=\textwidth]{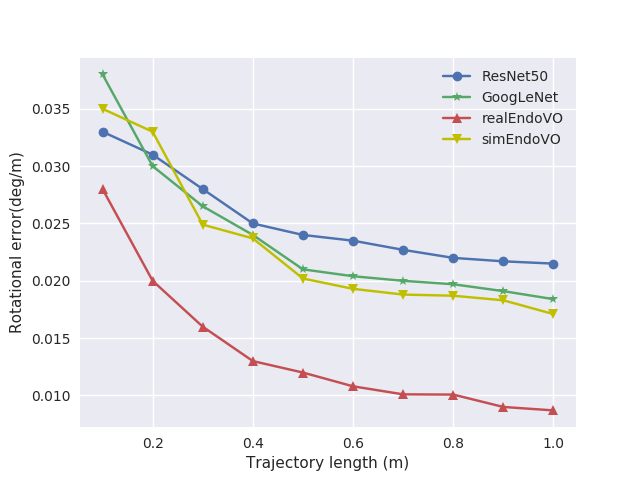}
\caption{Trajectory length vs totation error}
\label{fig:rot_error}       % Give a unique label
\end{subfigure}
%\\
~
\caption{Deep EndoVO outperforms both of the other models in terms of translational and rotational position estimation.}
\label{fig:error_vs_length}
\end{figure}

\begin{figure}[t!]
\begin{subfigure}[t]{0.5\textwidth} 
\includegraphics[width=\textwidth]{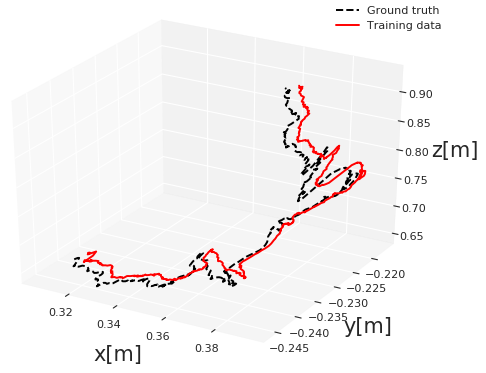}
\caption{Training good fitting}
\label{fig:goodfit_training} 
\end{subfigure}
~
\begin{subfigure}[t]{0.5\textwidth} 
\includegraphics[width=\textwidth]{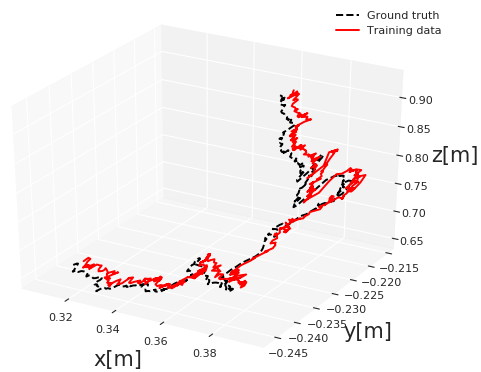}
\caption{Training overfitting}
\label{fig:overfit_training} 
\end{subfigure}
~
\begin{subfigure}[t]{0.5\textwidth} 
\includegraphics[width=\textwidth]{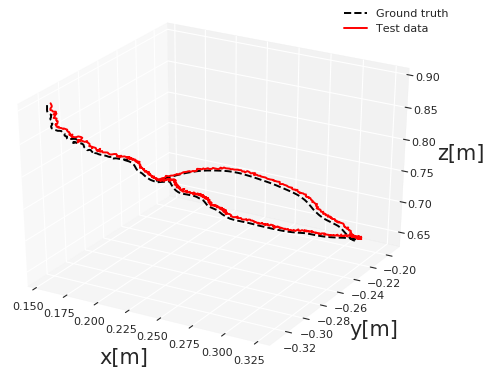}
\caption{Test good fitting}
\label{fig:goodfit_test}       % Give a unique label
\end{subfigure}
%\\
~
\begin{subfigure}[t]{0.5\textwidth} 
\includegraphics[width=\textwidth]{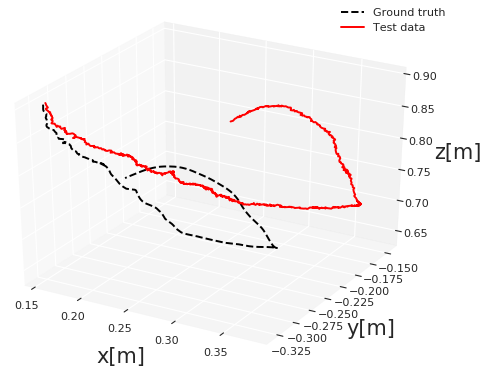}
\caption{Test overfitting}
\label{fig:overfit_test} 
\end{subfigure}
~

\caption{The affect of good fitting and overfitting. The first and the second rows show over-fitted and well-fitted models, respectively.  As seen in subfigures,
the model learns the details and noise in the training data to an undesired extent that it negatively impacts the performance of the model on the test data}
\label{fig:good_over_fit}
\end{figure}

Architecture was trained using Caffe library and NVIDIA Tesla K40 GPU. Using back-propagation-through-time method, the weights of hidden units were trained for up to $200$ epochs with an initial learning rate of $0.001$.  Overfitting meaning that the noise or random fluctuations in the training data are picked up and learned as concepts by the model, whereas these concepts do not apply to a new data and negatively affect the ability of the model to make generalizations,  was prevented using dropout and early stopping techniques (see Fig.\ref{fig:good_over_fit}). Dropout regularization technique introduced by \citep{srivastava2014dropout} is an extremely effective and simple method to avoid overfitting. It samples a part of the whole network and updates its parameters based on the input data. Early stopping is another widely used technique to prevent overfitting of a complex neural network architecture which was optimized by a gradient-based method. The approach is executed by splitting the dataset into a training and a validation set to evaluate the generalization capability of the model.

For the testing sessions, only real pig stomach recordings were used to ensure real world conditions. Additionally, we strictly avoided to use any frame from the training session for the testing session. Two separate experiments were conducted, whereas training session of the first experiment was performed using only the synthetic training dataset (see Fig.\ref{fig:037}) which we call simEndoVO and training session of the second experiment was performed using frames from both synthetic and real pig stomach dataset (see Fig. \ref{fig:037} and \ref{fig:07}) which we call realEndoVO. The performance of the simEndoVO and realEndoVO approaches were analysed using averaged Root Mean Square Errors (RMSEs) for translational and rotational motions. For various trajectories with different complexity levels of motions, including uncomplicated paths with slow incremental translations and rotations, comprehensive scans with many local loop closures and complex paths with sharp rotational and translational movements, we performed  testings on both simEndoVO and realEndoVO  comparing them with GoogLeNet and ResNet50 architectures which were modified to regress 6-DoF pose values by removing softmax layer and integrating a fully-connected (FC) layer and an affine regressor layer. The average translational and rotational RMSEs for simEndoVO, realEndoVO, GoogLeNet and ResNet50 networks against different path lengths are shown  in Fig. \ref{fig:error_vs_length}, respectively. The results depicted indicate, that realEndoVO clearly outperforms GoogLeNet and ResNet50, whereas simEndoVO slightly outperforms them. We presume that the effective use of LSTM in EndoVO architecture enabled learning motion dynamics across frame sequences, which is not feasible by architectures working with the principle of just-in-moment information processing; i.e GoogleNet and ResNet50. The results in Fig.\ref{fig:error_vs_length} also indicate that the training procedure including both simulator and real dataset was more informative than training only with simulator dataset. On the other hand, the accuracies achieved by the modified GoogLeNet are slightly better than accuracies achieved by the modified ResNet50, proving the superiority of inception layers over residual networks for feature extraction related tasks. Derived from RMSEs calculated, the rotational motion parameters seem to be more prone to overfitting compared to translational motion parameters (see Fig.\ref{fig:good_over_fit} for visual reference). The reason for that observation could be the fact that inner organ scanning procedures generally contain more translational motions than rotational motions resulting in a better learning for translations. As the length of the trajectory increases, both the translational and rotational error of all the proposed models significantly decrease (see Fig.\ref{fig:error_vs_length}). Some sample ground truth and estimated trajectories for realEndoVO, GoogLeNet and ResNet50 are shown in Fig.\ref{fig:trajectories} for visual reference. As seen in these sample trajectories, realEndoVO is able to stay close to the ground truth pose values for even sharp crispy motions, contrary to realEndoVO; GoogLeNet and ResNet50 path estimations which deviate drastically from the ground truth path values. Even for very fast and challenge paths such as \ref{fig:traj_1} and \ref{fig:traj_3}, the deviations of realEndoVO from the ground truth still remain in an acceptable range for medical operations. In addition to that, it is clearly seen that all of the three evaluated neural network architectures are able to estimate the scale very accurately without using any prior information or post alignment techniques contrary to traditional VO. Solving the scale ambiguity for monocular camera based VO makes our proposed DL based method more beneficial than traditional VO approach. As opposed to the traditional VO pipeline (see Fig.\ref{fig:trad}), the DL-based VO do not require any explicit feature extraction, matching, outlier detection or multi-scale bundle adjustment-like parameter tuning requiring operations, which can be seen as further benefits of the proposed approach.

\subsection{Comparisons of deep EndoVO with state-of-the-art SLAM methods}
\begin{figure}[t!]
\begin{subfigure}[t]{0.5\textwidth} 
\includegraphics[width=\textwidth]{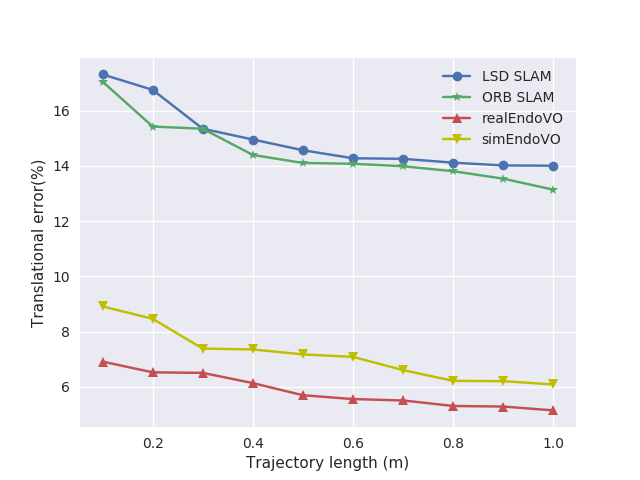}
\caption{Trajectory length vs translation error}
\label{fig:trans_error_trad} 
\end{subfigure}
~
\begin{subfigure}[t]{0.5\textwidth} 
\includegraphics[width=\textwidth]{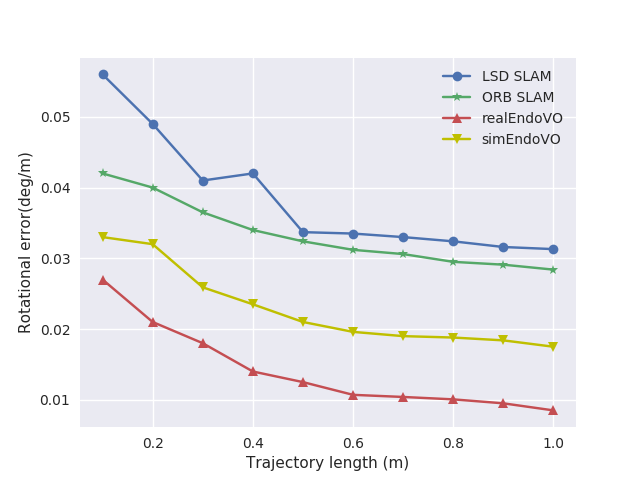}
\caption{Trajectory length vs rotation error}
\label{fig:rot_error_trad}       % Give a unique label
\end{subfigure}
%\\
~
\caption{Deep EndoVO outperforms the state-of-the-art SLAM methods ORB SLAM and LSD SLAM in both the translation and orientation estimation.}
\label{fig:error_vs_length_trad}
\end{figure}

\begin{figure}
\begin{subfigure}{0.5\textwidth} 
\includegraphics[width=\textwidth]{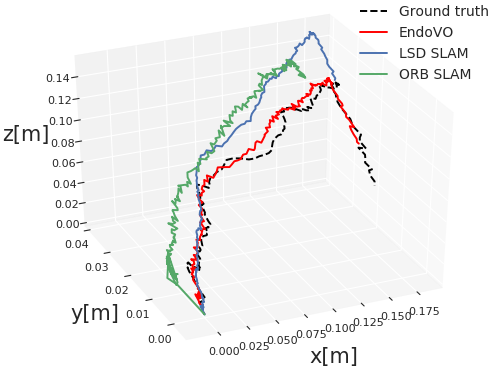}
\caption{Trajectory 1}
\label{fig:traj2_1}       % Give a unique label
\end{subfigure}
%~
\begin{subfigure}{0.5\textwidth} 
\includegraphics[width=\textwidth]{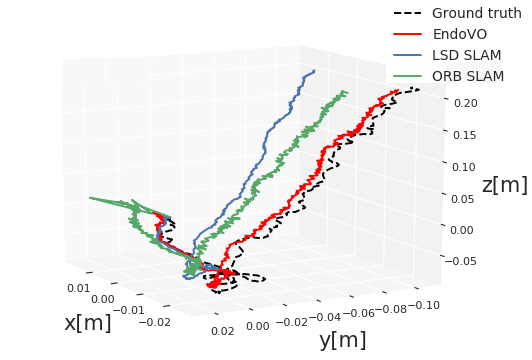}
\caption{Trajectory 2}
\label{fig:traj2_2} 
\end{subfigure}
%\\
\begin{subfigure}{0.5\textwidth} 
\includegraphics[width=\textwidth]{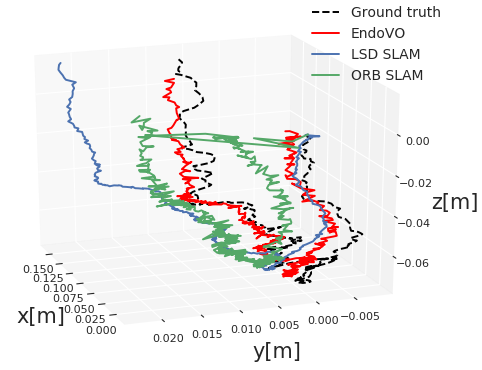}
\caption{Trajectory 3}
\label{fig:traj2_3} 
\end{subfigure}
%~
\begin{subfigure}{0.5\textwidth} 
\includegraphics[width=\textwidth]{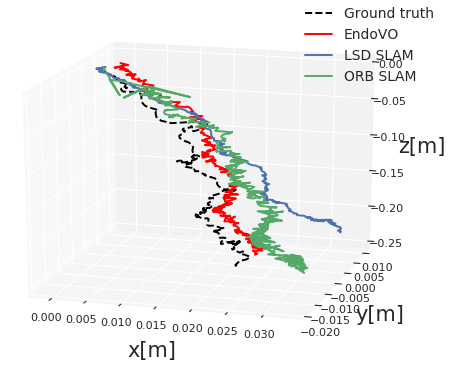}
\caption{Trajectory 4}
\label{fig:traj2_4} 
\end{subfigure}
%~
\caption{The ground truth and the trajectory plots acquired via deep EndoVO, LSD SLAM and ORB SLAM. Deep EndoVO is the closest to the ground truth trajectories compared to the state-of-the-art SLAM methods.}
\label{fig:trajectories_trad}
\end{figure}

In this subsection, we compare the performance of the proposed deep EndoVO with two of the widely used state-of-the-art SLAM methods; i.e. large-scale direct monocular SLAM (LSD SLAM) \citep{engel2014lsd} and the oriented fast and rotated brief SLAM (ORB SLAM) \citep{mur2015orb}. LSD SLAM is a direct image alignment-based method which optimizes the geometry using all of the image intensities. In addition to higher accuracy and robustness particularly in environments with little key points, this provides substantially more information about the geometry of the environment, which can be very valuable for medical robot applications, as well. ORB SLAM on the other hand, relies on feature point extraction and tracking to estimate camera pose and 3D map the environment. Even though it gives very promising results for feature-rich areas, its main deficiency appears once the robot enters poorly featured areas. Tracking failures are commonly observable for poorly featured GI tract tissues making ORB SLAM less proper for our case. We believe that our deep EndoVO architectures makes an optimal use of both direct and feature point information to estimate the pose. The average translational and rotational RMSEs for simEndoVO, realEndoVO, LSD SLAM and ORB SLAM, shown  in Fig. \ref{fig:error_vs_length_trad} indicate that both simEndoVO and realEndoVO clearly outperforms LSD SLAM and ORB SLAM in terms of pose accuracy. Sample trajectory estimations shown in Fig. \ref{fig:trajectories_trad} visualize clearly that the tracking capability of the proposed deep EndoVO is much more robust and reliable compared to LSD SLAM and ORB SLAM. In many parts of the trajectories, ORB SLAM and LSD SLAM deviate from the ground truth trajectory drastically, whereas deep EndoVO is still able to stay close to the ground truth values even for most challenge trajectory sections (see Fig.\ref{fig:traj2_2},\ref{fig:traj2_3}).

\section{CONCLUSION} \label{sec:conclusion}
In this study, we presented, to the best of our knowledge, the first deep VO method for endoscopic capsule robot and standard hand-held endoscope operations. The proposed system is able to achieve simultaneous representation learning and sequential modelling of motion dynamics across frames by concatenating the inception modules with RNN layers.  Many issues faced by traditional VO techniques such as feature correspondence establishment in low textured areas, high reflections, motion blur and low image quality are handled by the proposed deep EndoVO successfully. Since it is trained in an end-to-end manner, there is no need to carefully fine-tune the parameters of the system.  As a future step, we consider to combine deep EndoVO with some functionalities from the traditional VO pipelines such as RANSAC for outlier detection and bundle fusion for globally consistent pose estimation etc to avoid drifts. Moreover, we consider to develop a stereo version of the proposed deep EndoVO approach.

%\begin{figure}
% Use the relevant command to insert your figure file.
% For example, with the graphicx package use
%\includegraphics[width=0.75\textwidth]{figures/fig_rot_err.png}
% figure caption is below the figure
%\caption{As the training proceeds, the rotation error decreases.}
%\label{fig:rot_err}       % Give a unique label
%\end{figure}

\pagebreak

\section*{References}

\bibliography{mybibfile}

\end{document}